\documentclass[11pt,a4paper]{article}
\usepackage[hyperref]{acl2018}
\usepackage{times}
\usepackage{latexsym}
\usepackage{graphicx}
\usepackage{algorithm,algpseudocode}
\usepackage{xcolor}

\usepackage{url}

\aclfinalcopy 

\title{Nearly Zero-Shot Learning for Semantic Decoding in Spoken Dialogue Systems}

\author{Lina M.~Rojas-Barahona, Stefan~Ultes, Pawel Budzianowski, \\
{\bf I\~{n}igo Casanueva, Milica~Ga{\v s}i{\' c}, Bo-Hsiang Tseng and Steve~Young}\\
  Department of Engineering, University of Cambridge, Cambridge, UK\\
  {lmr46, mg436, sjy}@cam.ac.uk}

\date{}

\begin{document}
\maketitle
\begin{abstract}
  This paper presents two ways of dealing with scarce data in semantic decoding using N-Best speech recognition hypotheses. First, we learn features by using a deep learning architecture in which the weights for the unknown and known categories are jointly optimised. Second, an unsupervised method is used for further tuning the weights.  Sharing weights injects prior knowledge to unknown categories. The unsupervised tuning (i.e. the risk minimisation) improves the  F-Measure when recognising nearly zero-shot data on the DSTC3 corpus. This unsupervised method can be applied subject to two assumptions: the rank of the class marginal is assumed to be known and the class-conditional scores of the classifier are assumed to follow a Gaussian distribution.  
\end{abstract}

\section{Introduction}

The semantic decoder in a dialogue system is the component in charge of processing the automatic speech recognition (ASR) output and predicting the semantic representation. In slot-filling dialogue systems, the semantic representation consists of a dialogue act and a set of slot-value pairs. For instance, the semantic representation of the utterance "uhm I am looking for a restaurant in the north of town" will have the semantics: inform(type=restaurant,area=north), where \textit{inform} is the dialogue act, \textit{type} and \textit{area} are slots and \textit{restaurant} and \textit{north} are their respective values. 
Making the semantic decoder robust to rare slots is a crucial step towards open-domain language understanding.

In this paper, we deal with rarely seen slots by following two steps. (i) We optimise jointly in a deep neural network the weights that feed multiple binary Softmax units.  (ii) We further tune the weights learned in the previous step by minimising the theoretical risk of the binary classifiers as proposed in~\cite{riskminim}. In order to apply the second step, we rely on two assumptions:
the rank of the class marginal is assumed to be known and the class-conditional linear scores are assumed to follow a Gaussian distribution. In ~\cite{riskminim}, this approach has been proven to converge towards the true optimal classifier risk.
We conducted experiments on the dialogue corpus released for the third dialogue state tracking challenge, namely DSTC3~\cite{thethirdstc3} and we show positive results for detecting rare slots as well as zero-shot slot-value pairs.

\section{Related Work}
\let\center\flushleft
\let\endcenter\endflushleft
\maketitle

Previous work on  domain adaptation improved discriminative models by using \textit{priors} and \textit{feature augmentation}~\cite{daume2009frustratingly}. The former uses the weights of the classifier in the known domain as a prior for the unknown domain. The latter extends the feature space with general features that might be common to both domains.  


Recently, feature-based adaptation has been refined with unsupervised auto-encoders that learn features that can generalise between domains~\cite{glorot2011domain,zhou2016bi}. These models have proven to be successful for sentiment analysis but not for more complex semantic representations.
A popular way to support semantic generalisation is to use high dimensional word vectors trained on a very large amount of data~\cite{mikolov2013efficient, pennington2014glove} or even cross-lingual data~\cite{TACL1171}.

Previous approaches for recognising scarce slots in spoken language understanding relied on the semantic web~\cite{tur2012exploiting},  linguistic resources~\cite{gardent:hal-00905405},  open domain knowledge bases (e.g.,  NELL, freebase.com)~\cite{pappu2013predicting}, user feedback~\cite{ferreira2015online} or generation of synthetic data by simulating ASR errors~\cite{zhu2014semantic}. 

Unlike most of the state-of-the-art models~\cite{liu2016joint,mesnil2013investigation}, in this work semantic decoding is not treated as a sequence model because of the lack word-aligned semantic annotations.  In this paper, we inject priors as proposed in~\cite{daume2009frustratingly}. Moreover, our work differs from his because the priors are given by the weights trained through a joint optimisation of several binary Softmax units within a deep architecture exploiting word vectors. In this way, the rare slots exploit the embedded information learned about the known slots. Furthermore, we propose an unsupervised method for further tuning the weights by minimising the theoretical risk. 

\section{Deep Learning Semantic Decoder}
\label{s:dlsemi}

\begin{figure}[ht]
\centering
\includegraphics[scale=0.3]{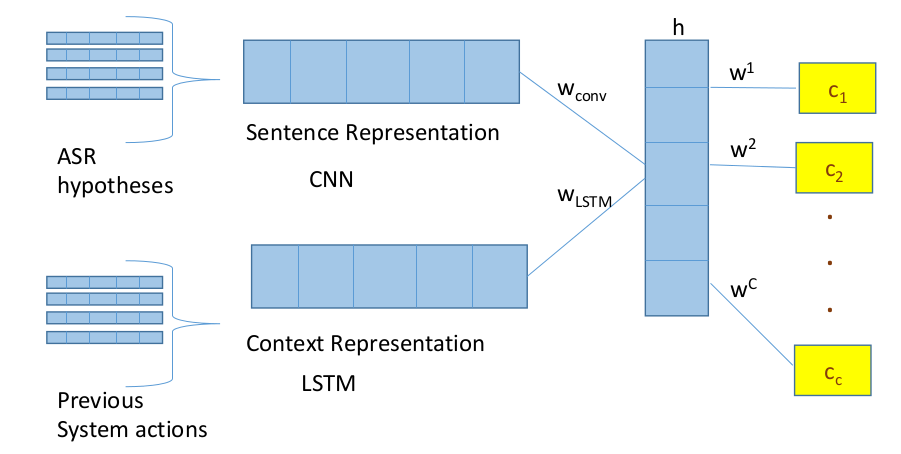}
\caption{Combination of sentence and context representations for the joint optimisation. In the first step, the binary classifiers $c_i$ are predicting the presence or absence of slots. In the second step, they are predicting the presence or absence of the values for a given slot.}
\label{f:dlarchi}
\end{figure}

The Deep Learning semantic decoder is similar to the one proposed in~\cite{barahona2016exploiting}. It has been split into two steps: (i) detecting the slots and (ii) predicting the values per slots. The deep architecture depicted in Figure~\ref{f:dlarchi} is used in both steps. It combines sentence and context representations, applying a non linear function to their weighted sum (Eq.\ref{eq:comb}), to generate the final hidden unit that feeds various binary Softmax outputs (Eq.\ref{eq:softmax}).  

\begin{equation}
   \mathbf{h}=tanh(\mathbf{W_{conv}}\cdot\mathbf{sent}+\mathbf{W_{LSTM}}\cdot\mathbf{ctxt})
   \label{eq:comb}
\end{equation}
\begin{equation}
P(Y=k|\mathbf{h},\mathbf{W}) = \frac{e^{(\mathbf{W}_{k}\mathbf{h})}}{\sum_{k\prime} e^{(\mathbf{W}_{k\prime}\mathbf{h})}}
 \label{eq:softmax}
\end{equation}
where $k\in\{0,1\}$ is the index of the output neuron representing one class.

The sentence representation ($\mathbf{sent}$) is obtained through a convolutional neural network (CNN) that processes the 10 best ASR hypotheses. The context representation ($\mathbf{ctxt}$) is a long short-term memory (LSTM) that has been trained with the previous system dialogue acts. 
In the first step, there are as many Softmax units as slots (Figure~\ref{f:dlarchi}).  In the second step, a distinct model is trained for each slot and there are as many distinct Softmax units as possible values per slot (i.e. as define by an ontology ). For instance, the model that predicts food, will have $86$ Softmax units. One that predicts the presence or absence of "Italian" food, another unit  that predicts "Chinese" food and so on. 

All the weights in the neural network are optimised jointly. The benefits of joint inference have been published in the past for different NLP tasks~\cite{Singh2013,liu2016joint}. The main advantage of joint-inference is that parameters are shared between predictors, thus weights can be adjusted based on their mutual influence. For instance, the most frequent slots might influence infrequent slots.

\section{Risk Minimisation (RM)}
\label{s:rm}

We use the unsupervised approach proposed in~\cite{riskminim} for risk minimisation (RM). 
We assume a binary classifier that associates a score $f_{\mathbf{W}_0}(\mathbf{h})$ to the first class 0 for the hidden unit $\mathbf{h}=(h_1,\cdots,h_{n})$ of dimension $n$:
$$f_{W_0}(\mathbf{h}) = \sum_i^{n} w_i h_i$$
where the parameter $w_i \in {\rm I\!R}$ represents the weight of the feature indexed by $i$ for class 0.

The objective of training is to minimize the classifier risk:
\begin{equation}\label{eq:r}
R(\mathbf{W})=E_{p(\mathbf{h},Y)}[\mathcal{L}(Y,f_\mathbf{W}(\mathbf{h}))]
\end{equation}
where $Y$ is the true label 
and $\mathcal{L}(Y,f_\mathbf{W}(\mathbf{h}))$ is the loss function.
The risk is derived as follows:
{\small{
\begin{equation}\label{eq:rr}
R(\mathbf{W})=\sum_{y\in\{0,1\}} P(y) \int_{-\infty}^{+\infty} P(f_\mathbf{W}(\mathbf{h})=\alpha|y)\mathcal{L}(y,\alpha)d\alpha
\end{equation}
}}

We use the following hinge loss:
\begin{equation}\label{eq:loss}
\mathcal{L}(y,\alpha) = \left( 1+\alpha_{1-y} - \alpha_y\right)_+
\end{equation}
where $(z)_+ = \max (0,z)$, and $\alpha_y=f_{\mathbf{W}_y}(\mathbf{h})$ is the linear score for the correct class $y$.
Similarly, $\alpha_{1-y}=f_{\mathbf{W}_{1-y}}(\mathbf{h})$ is the linear score for the wrong class.

Given $y$ and $\alpha$, the loss value in the integral (Equation~\ref{eq:rr}) can be computed easily. Two terms remain: $P(y)$ and $P(f_\mathbf{W}(\mathbf{h})=\alpha|y)$.
The former is the class marginal and is assumed to be known.
The latter is the class-conditional distribution of the linear scores, which is assumed to be normally distributed.
This implies that
$P(f_\mathbf{W}(\mathbf{h}))$ is distributed as a mixture of two Gaussians (GMM):
$$P(f_\mathbf{W}(\mathbf{h})) = \sum_{y\in\{0,1\}} P(y)\mathcal{N}(f_\mathbf{W}(\mathbf{h});\mu_y,\sigma_y)$$
where
$\mathcal{N}(z;\mu,\sigma)$
is the normal probability density function.
The parameters $(\mu_0,\sigma_0,\mu_1,\sigma_1)$ can be estimated
from an \textbf{unlabeled corpus} $\mathcal{U}$ using a standard Expectation-Maximization (EM) algorithm for GMM training.
Once these parameters are known, it is possible to compute the integral in Eq.~\ref{eq:rr} and thus an estimate $\hat R(\mathbf{W})$ of the risk
without relying on any labeled corpus. In ~\cite{riskminim}, it has been proven that: (i) the Gaussian parameters estimated with EM converge towards their true values, (ii) $\hat R(\mathbf{W})$ converges towards the true risk $R(\mathbf{W})$ and (iii) the estimated optimum converges towards the true optimal parameters, when the size of the unlabeled corpus increases infinitely. This is still true even when the class priors $P(y)$ are unknown.

The unsupervised algorithm is as follows:
\paragraph{\scriptsize{Unsupervised tuning for the binary classifier $c$, where $c=1,...,C$}}
{\small{
\begin{algorithmic}[1]\raggedright
\scriptsize{
\State \textbf{input:}  $\mathbf{h}$  the top hidden layer and the weights $\mathbf{W}^c$, as trained by the  deep learning decoder (Section~\ref{s:dlsemi}). 
\State \textbf{output:} The tuned weights $\hat \mathbf{W}^c$
\Repeat

\For{every index $i$ in $\mathbf{h}$} , 
\State Change the weights $\mathbf{W}^c_i=\mathbf{W}^c_i+\delta$, 
\State Estimate the Gaussian parameters using EM 
\State Compute the risk (Eq.~\ref{eq:rr})\footnote{A closed-form is used to compute the risk for binary classifiers.~\cite{rojasbarahona:hal-01184849}} on the unlabeled corpus $\mathcal{U}$ (i.e. the evaluation set). 
\State Compute the gradient using finite differences
\State Update the weights  accordingly $\hat \mathbf{W}^c_i=\mathbf{W}^c_i$
\EndFor
\Until{convergence}
}
\end{algorithmic}
}
}


\section{Experiments}
The supervised and unsupervised models are evaluated on  DSTC3~\cite{thethirdstc3} using the macro F-Measure\footnote{The macro F-score was chosen because we are evaluating the capacity of the classifiers to predict the correct class and both classes positive and negative are equally important for our task. Moreover, being nearly zero-shot classifiers, it would be unfair to evaluate only the capacity of predicting the positive category.}. 
We compare then three distinct models, (i) independent neural models for every binary classifier; (ii) neural models optimised jointly and (iii) further tuning of the weights through RM.

\paragraph{Dataset}
As displayed in Table~\ref{t:dstc3} in DSTC3 new slots were introduced relative to DSTC2. The training set contains only a few examples of these slots while the test set contains a large number of them.  Interestingly, frequent values per slots in the trainset such as \textit{area=north}, are absolutely absent in the testset. 
In DSTC3 the dialogues are related to restaurants, pubs and coffee shops. 
The new slots are: \textit{childrenallowed}, \textit{hastv}, \textit{hasinternet} and \textit{near}. Known slots, such as \textit{food}, can have zero-shot values as shown in Table~\ref{t:dstc3vals}. The corpus contains $3246$ dialogues, $25610$ turns in the trainset and $2264$ dialogues, $18715$ turns in the testset.

\paragraph{Hyperparameters and Training}
The neural models were implemented in Theano~\cite{bastien2012theano}. 
We used filter windows of 3, 4, and 5 with 100 feature maps each for the CNN. A dropout rate of $0.5$ and a batch size of 50 was employed. Training is done through stochastic gradient descent over shuffled mini-batches with the Adadelta update rule. 
GloVE word vectors were used~\cite{pennington2014glove} to intialise the models with a dimension $n=100$. 
For the context representation, we use a window of the 4 previous system acts. 
The risk minimisation gradient descent runs during 2000 iterations for each binary classifier and the class priors $P(y)$ were set to $0.01$ and $0.99\%$ for the positive and negative classes respectively.

\begin{table}[htbp]
\centering
  \begin{tabular}{l | ll }
 \hline
 \scriptsize{Slot}&\scriptsize{$\#$Train}&\scriptsize{$\#$Test}\\
 \hline
 \scriptsize{\textbf{hastv}}&\scriptsize{1}&\scriptsize{239}\\
 \scriptsize{\textbf{childrenallowed}}&\scriptsize{2}&\scriptsize{119}\\
 \scriptsize{\textbf{near}}&\scriptsize{3}&\scriptsize{74}\\
 \scriptsize{\textbf{hasinternet}}&\scriptsize{4}&\scriptsize{215}\\
 \scriptsize{area}&\scriptsize{3149}&\scriptsize{5384}\\
 \scriptsize{food}&\scriptsize{5744}&\scriptsize{7809}\\
 \hline
 \end{tabular}
 \caption{Frequency of slots in DSTC3.}
 \label{t:dstc3}
 \end{table}
 
\begin{table}[htbp]
\centering
  \begin{tabular}{ll | ll }
 \hline
 \scriptsize{Slot}&\scriptsize{Value}&\scriptsize{$\#$Train}&\scriptsize{$\#$Test}\\
 \hline 
 \scriptsize{near}&\scriptsize{trinity college}&\scriptsize{0}&\scriptsize{5}\\
 \scriptsize{food}&\scriptsize{american}&\scriptsize{0}&\scriptsize{90}\\
  \scriptsize{food}&\scriptsize{chinese takeaway}&\scriptsize{0}&\scriptsize{87}\\
   \scriptsize{area}&\scriptsize{romsey}&\scriptsize{0}&\scriptsize{127}\\
   \scriptsize{area}&\scriptsize{girton}&\scriptsize{0}&\scriptsize{118}\\
   
 \hline
 \end{tabular}
 \caption{Some zero-shot values per slots in DSTC3.}
 \label{t:dstc3vals}
 \end{table}

 \paragraph{The Gaussianity Assumption}
 As explained in Section~\ref{s:rm}, the risk minimisation tuning assumes the class-conditional linear scores are distributed normally. We verified this assumption empirically on our unlabeled corpus $\mathcal{U}$ (i.e. DSTC3 testset) and we found that for the slots: \textit{childrenallowed}, \textit{hastv} and \textit{hasinternet} this assumption holds. However, the distribution for \textit{near} has a negative skew. When verifying the values per slot, this assumption does not hold for \textit{area}. Therefore, we can not guarantee this method will work correctly for \textit{area} values on this evaluation set. 
 


 \begin{table}
 \centering{
 \begin{tabular}{l|l}
 \hline
 \multicolumn{2}{c}{\scriptsize{Deep Learning Independent Models}}\\\hline
 \scriptsize{Slot}&
 \scriptsize{F-Measure}\\\hline
 \scriptsize{childrenallowed}
 &\scriptsize{$49.84$\%}\\
 \scriptsize{hastv}&
 \scriptsize{$49.68$\%}\\
 \scriptsize{hasinternet}&
\scriptsize{$49.72$\%}\\
 \scriptsize{near}&
 \scriptsize{$49.90$\%}\\\hline
\multicolumn{2}{c}{\scriptsize{Deep Learning Joint Optimisation}}\\\hline
 \scriptsize{childrenallowed}&
 \scriptsize{$58.76$\%}\\
 \scriptsize{hastv}&
 \scriptsize{$59.16$\%}\\
 \scriptsize{hasinternet}&
 \scriptsize{$58.77$\%}\\
 \scriptsize{near}&
 \scriptsize{$56.65$\%}\\
 \hline
\multicolumn{2}{c}{\scriptsize{Risk Minimisation Tuning}}\\\hline
 \scriptsize{childrenallowed}&
 \scriptsize{$\mathbf{61.64}$\%}\\
 \scriptsize{hastv}&
 \scriptsize{$\mathbf{61.35}$\%}\\
 \scriptsize{hasinternet}&
 \scriptsize{$\mathbf{60.87}$\%}\\
 \scriptsize{near}&
 \scriptsize{$\mathbf{58.60}$\%}\\\hline  
 \end{tabular}}
 \caption{Results for learning rare slots on DSTC3 evaluation set.}
 \label{t:dlsemisres}
 \end{table}

 \begin{table}[ht!]
 \centering{
 \begin{tabular}{ll|l}
 \hline
 \multicolumn{3}{c}{\scriptsize{Deep Learning Independent Models}}\\\hline
 \scriptsize{Slot}&\scriptsize{Value}&
 \scriptsize{F-Measure}\\\hline
 \scriptsize{near}&\scriptsize{trinity college}&
 \scriptsize{$49.99$\%}\\
 \scriptsize{food}&\scriptsize{american}&
 \scriptsize{$49.88$\%}\\
 &\scriptsize{chinese take away}&
 \scriptsize{$49.88$\%}\\
 \scriptsize{area}&\scriptsize{romsey}&
 \scriptsize{$49.83$\%}\\
 &\scriptsize{girton}&
 \scriptsize{$49.84$\%}\\
 \hline
\multicolumn{3}{c}{\scriptsize{Deep Learning Joint Optimisation}}\\\hline
\scriptsize{near}&\scriptsize{trinity college}&
\scriptsize{$61.25$\%}\\
 \scriptsize{food}&\scriptsize{american}&
 \scriptsize{$59.93$\%}\\
 &\scriptsize{chinese take away}&
 \scriptsize{$61.02$\%}\\
 \scriptsize{area}&\scriptsize{romsey}&
 \scriptsize{$\mathbf{51.30}$\%}\\
 &\scriptsize{girton}&
 \scriptsize{$\mathbf{55.19}$\%}\\
 \hline
\multicolumn{3}{c}{\scriptsize{Risk Minimisation Tuning}}\\\hline
\scriptsize{near}&\scriptsize{trinity college}&
\scriptsize{$\mathbf{62.08}$\%}\\
 \scriptsize{food}&\scriptsize{american}&
 \scriptsize{$\mathbf{62.52}$\%}\\
 &\scriptsize{chinese take away}&
 \scriptsize{$\mathbf{63.79}$\%}\\
 \scriptsize{area}&\scriptsize{romsey}&
 \scriptsize{$48.76$\%}\\
 &\scriptsize{girton}&
 \scriptsize{$51.45$\%}\\
 \hline
 \end{tabular}}
 \caption{Results for learning zero shot slot-value pairs on DSTC3 evaluation set.}
  \label{t:dlsemisvres}
 \end{table}

\section{Results}
Tables~\ref{t:dlsemisres} and \ref{t:dlsemisvres} display the performance of the models that predict slots and values respectively.  The low F-Measure in the independent models shown their inability to predict positive examples. The models improve significantly the precision and F-Measure after the joint-optimisation. Applying RM tuning results in the best F-Measure for all the rare slots (Table~\ref{t:dlsemisres}) and for the values of the slots \textit{food} and \textit{near} (Table~\ref{t:dlsemisvres}).
For \textit{area}, the joint optimisation improves the F-Measure but the improvement is lower than for other slots. The performance is being affected by its low cardinality (i.e. $20$), the high variability of new places and the fact that frequent values such as  \textit{north} and  \textit{east}, are completely absent in the test set. As suspected,  the RM tuning degraded the precision and F-Measure because the Gaussianity assumption does not hold for \textit{area}. However, RM will work well in larger evaluation sets because the Gaussian assumption will hold when the unlabeled corpus tends to infinite (please refer to ~\cite{riskminim} for the theoretical proofs).


\section{Conclusion}
We presented here two novel methods for zero-shot learning in a deep semantic decoder. First, features and weights were learned through a joint optimisation within a deep learning architecture. Second, the weights were further tuned through risk minimisation. We have shown that the joint optimisation significantly improves the neural models for nearly zero-shot slots. We have also shown that under the Gaussianity assumption, the RM tuning is a promising method for further tuning the weights of zero-shot data in an unsupervised way.

\bibliographystyle{acl_natbib}
\bibliography{naaclhlt2018}

\end{document}